\documentclass[runningheads]{llncs}
\usepackage{subcaption}
\usepackage{graphicx}
\usepackage{comment}
\usepackage{amsmath,amssymb} 
\usepackage{color}

\usepackage[width=122mm,left=12mm,paperwidth=146mm,height=193mm,top=12mm,paperheight=217mm]{geometry}
\usepackage{caption}
\usepackage{subcaption}
\usepackage{multirow}

\begin{document}
	
\pagestyle{headings}
\mainmatter
\def\ECCVSubNumber{3202}  

\title{Distillation Guided Residual Learning for Binary Convolutional Neural Networks} 


\author{Jianming Ye, Shiliang Zhang, Jingdong Wang}
\institute{jmye@pku.edu.cn, slzhang.jdl@pku.edu.cn, jingdw@microsoft.com}

\maketitle

\begin{abstract}
It is challenging to bridge the performance gap between Binary CNN (BCNN) and Floating point CNN (FCNN). We observe that, this performance gap leads to substantial residuals between intermediate feature maps of BCNN and FCNN. To minimize the performance gap, we enforce BCNN to produce similar intermediate feature maps with the ones of FCNN. This training strategy, \emph{i.e.}, optimizing each binary convolutional block with block-wise distillation loss derived from FCNN, leads to a more effective optimization to BCNN. It also motivates us to update the binary convolutional block architecture to facilitate the optimization of block-wise distillation loss. Specifically, a lightweight shortcut branch is inserted into each binary convolutional block to complement residuals at each block. Benefited from its Squeeze-and-Interaction (SI) structure, this shortcut branch introduces a fraction of parameters, \emph{e.g.}, 10\% overheads, but effectively complements the residuals. Extensive experiments on ImageNet demonstrate the superior performance of our method in both classification efficiency and accuracy, \emph{e.g.}, BCNN trained with our methods achieves the accuracy of 60.45\% on ImageNet.
\end{abstract}

\section{Introduction}

Many milestone works~\cite{VGG,krizhevsky2012alexnet,he2016resnet} have been conducted to design deeper and more powerful Convolutional Neural Network (CNN) architectures. Thanks to those efforts, the performance of deep CNNs has been significantly boosted. Meanwhile, existing deep CNNs usually consist of millions of parameters and consume billions of Floating Point Operations Per second (FLOPs) for computation. Those properties limit their applications in scenarios with limited computation and memory capabilities. As there are growing demands to run vision tasks on portable devices, many research efforts~\cite{zhang2018shufflenet,Liu_2019_ICCV,liu2019auto,zhou2016dorefa} aim to reduce the space and computational complexities. One popular strategy is to convert Floating point CNNs (FCNNs) into Binary CNNs (BCNNs), where the binarization significantly improves the computation and memory efficiency.

\begin{figure}
\centering
		\includegraphics[width=0.75\linewidth]{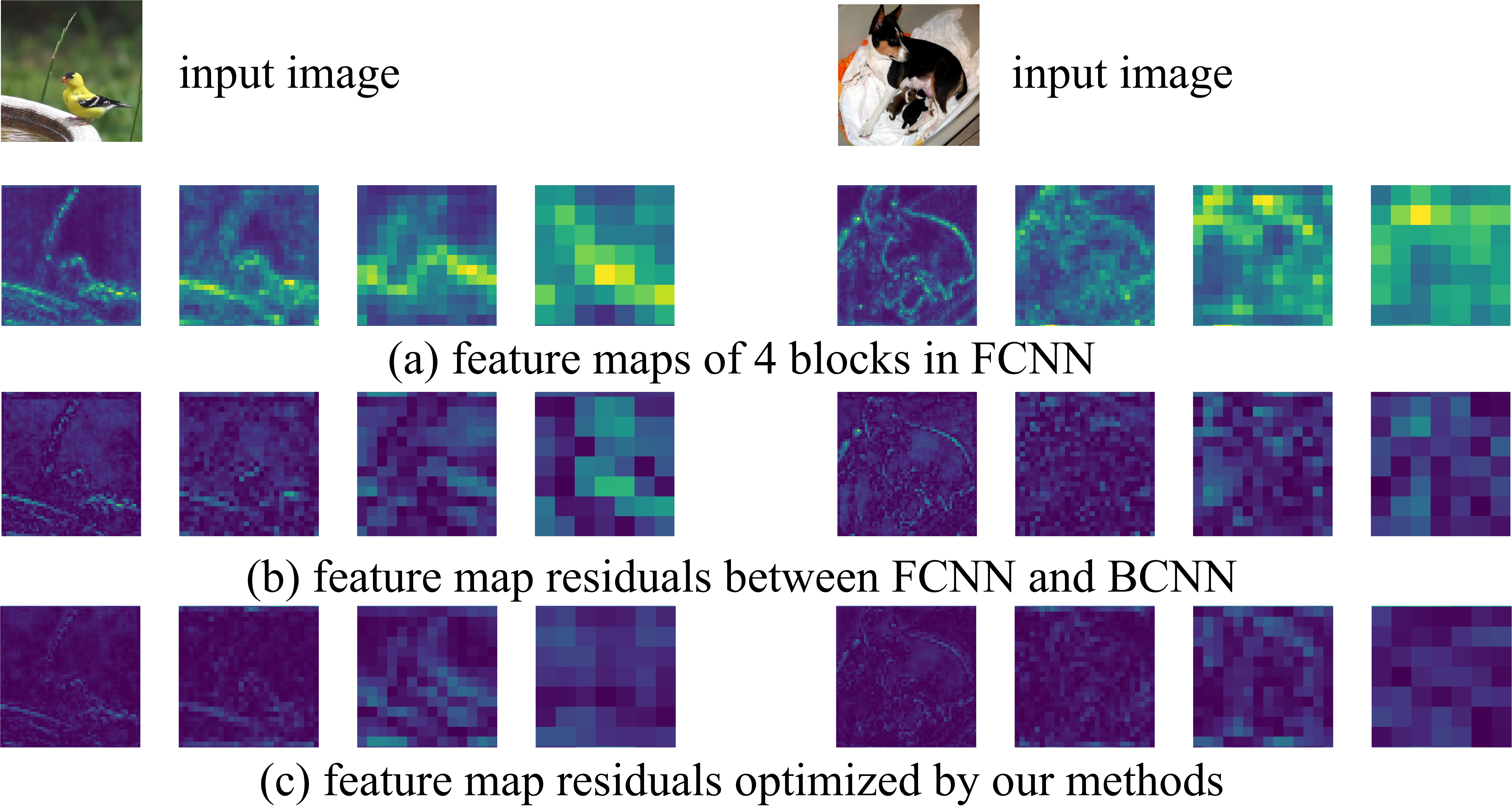}

	\caption{Visualization of intermediate feature maps from 4 convolutional blocks of ResNet18~\cite{he2016resnet} in (a) and feature map residuals between ResNet18 and baseline BCNN~\cite{bethge2019Simplicity} in (b). (c) illustrates feature map residuals optimized by our methods.}
	\label{fig:feamap}

\end{figure}
As an early BCNN work, BNN~\cite{BNN} is proposed to train networks with weights and activations constrained to $\pm 1$. It is efficient but exhibits degraded performance on large dataset like ImageNet~\cite{deng2009imagenet}.	Many recent works like XNOR-Net~\cite{rastegari2016xnor}, ABC-Net~\cite{lin2017ABC-Net}, Bi-Real Net~\cite{liu2018bireal}, PCNN~\cite{gu2019projection}, BONNs~\cite{gu2019bayesian} and CI-BCNN~\cite{wang2019CI-BCNN} have continuously boosted the performance of BCNN, \emph{e.g.}, CI-BCNN~\cite{wang2019CI-BCNN}, achieves ImageNet classification accuracy of 59.9\%, substantially better than the 51.2\% of XNOR-Net~\cite{rastegari2016xnor}. However, there still exists a substantial performance gap between BCNNs and FCNNs, which easily achieve accuracy of 69\% on ImageNet~\cite{deng2009imagenet}. More detailed review to BCNNs will be given in Sec.~\ref{sec:relatedwork}.

Compared with FCNN, BCNN shows limited modeling capability because of its binary convolutional kernels. Meanwhile, BCNN training is not as efficient as the training of FCNN. For instance, it is difficult to implement gradient back prorogation with binary parameters. Therefore, BCNN training involves two sets of parameters~\cite{BNN}, \emph{i.e.}, binarized parameters and floating point parameters, respectively. Binarized parameters are used for forward propagation and loss computation. Floating point parameters are used for loss back propagation and parameter updating. Inferior convolutional layers and training strategy result in substantially different network responses. As shown in Fig.~\ref{fig:feamap} (a) and (b), FCNN and BCNN generate different intermediate feature maps for the same input. Such block-wise residuals may accumulate as the network goes deeper, resulting in substantial performance gap at the output block.

This work targets to study more efficient BCNN training strategies, as well as more effective BCNN architectures. Training BCNNs with back-propagation suffers from vanishing gradient and quantization error in parameter binarization. To alleviate those issues, we leverage intermediate feature maps of FCNN for BCNN training. This can be implemented by training each binary convolutional block with distillation loss. In other words, each BCNN block is supervised to produce similar outputs with its corresponding FCNN block. Existing methods mostly use FCNN for BCNN initialization. Compared with those works, block-wise distillation loss could better leverage FCNN in BCNN training and shows potential to eliminate aggregated residuals as network goes deeper.

Limited model capability of binary convolution hinders BCNN to simulate the behavior of FCNN. This makes it hard to optimize block-wise residuals as shown in Fig.~\ref{fig:feamap}. We propose to complement the residuals with additional shortcut branches, which are inserted into each binary convolutional block to enhance model capability. Compared with original feature maps, residual feature maps exhibit limited variance. Therefore, we manage to model them with a lightweight Squeeze-and-Interaction (SI) shortcut. Specifically, to compute a residual feature map with $C$ channels, the squeeze module first computes a feature map with $S, S<C$ channels, which is then feed into the interaction module to recover feature map with $C$ channels. The parameter $S$ is block-dependent and is automatically selected. SI shortcut introduces marginal parameter overheads, but facilitates optimization of block-wise distillation loss.

We conduct image classification experiments on two widely used datasets CIFAR-10~\cite{cifar10} and ImageNet~\cite{deng2009imagenet}. Experimental results show that our distillation loss and SI shortcut effectively boost the BCNN performance. As illustrated in Fig.~\ref{fig:feamap} (c), feature maps from BCNN optimized by our method are more similar to the ones by FCNN. It is also interesting to observe that,
SI 
shortcuts only take about 10\% parameter overheads, but substantially boosts the classification accuracy, \emph{e.g.}, we achieve 60.45\% accuracy on ImageNet. We also make comparisons with other recent BCNNs, where our method achieves competitive performance, \emph{e.g.}, 2.13\% better than the recent IR-Net~\cite{qin2020forward} in accuracy on ImageNet.

Most of current works use FCNNs for BCNN initialization. To the best of our knowledge, this is an early work leveraging FCNNs for BCNN optimization through block-wise knowledge distillation. Another recent work~\cite{liu2019ganbin} leverages FCNN in BCNN training through training GANs. Compared with this work, our model performs better and does not need to train GANs, hence could be easier to train and optimize. Our promising performance also thanks to the introduction of SI shortcuts, which complement the block-wise residuals, thus facilitate the BCNN optimization. Those novel components guarantee the competitive performance of our methods.

\section{Related Work} \label{sec:relatedwork}

Existing works on Binary Convolutional Neural Network (BCNN) can be roughly summarized into two categories according to their parameter overheads.

The first category improves the performance of BCNN by introducing new convolutional layers, loss functions~\cite{BNN,rastegari2016xnor,liu2018bireal,MoBiNet,liu2019ganbin,gu2019bayesian,wang2019CI-BCNN}, \emph{etc}. BNN~\cite{BNN} achieves high classification accuracy on small datasets, like CIFAR-10~\cite{cifar10}, but does not perform well on large-scale ImageNet~\cite{deng2009imagenet}. XNOR-Net~\cite{rastegari2016xnor} boosts performance by introducing binary convolutional kernels with scalars.
Bi-Real Net~\cite{liu2018bireal} further enhances performance by connecting real activations before sign function of the next block. RBCN~\cite{liu2019ganbin} trains GANs to affiliate BCNN training. BONNs~\cite{gu2019bayesian} develops a novel Bayesian learning algorithm.
The recent CI-BCNN~\cite{wang2019CI-BCNN} mines channel-wise interactions through reinforcement learning.

The other category introduces more parameters~\cite{lin2017ABC-Net,Zhuang2019SBNN,gu2019projection,Liu2019CBCN,zhu2019BENN}, and achieves better performance.
Related works include ABC-Net~\cite{lin2017ABC-Net}, GroupNet~\cite{Zhuang2019SBNN}, PCNN~\cite{gu2019projection}, CBCN~\cite{Liu2019CBCN} and BENN~\cite{zhu2019BENN}.
Both ABC-Net~\cite{lin2017ABC-Net} and GroupNet~\cite{Zhuang2019SBNN} fuse several binary convolutional layers to approximate one floating point convolutional layer. PCNN~\cite{gu2019projection} learns a set of diverse quantized kernels to improve the performance. CBCN~\cite{Liu2019CBCN} proposes a circulant binary convolution, which takes about 16 times more calculations than a simple BCNN. BENN~\cite{zhu2019BENN} regards BCNNs as weak classifiers and uses AdaBoost~\cite{freund1997decision} to ensemble these classifiers.

This work belongs to the first category. Our work differs with previous ones in that, it uses block-wise distillation loss to train each binary convolutional block. Some works leverage knowledge distillation to train quantized networks~\cite{mishra2017apprentice,polino2018model,zhuang2018towards}. However, training binary networks with knowledge distillation is still under-explored. The reason could be that, the limited model capability of BCNN makes it hard to simulate the real-value network, leading to difficulty in training convergence. Our SI shortcut boosts the model capability and simplifies the block-wise distillation loss optimization. MoBiNet~\cite{MoBiNet} adds parameter-free shortcuts between CNN blocks to prevent vanishing gradient and make convergence easier. Different from MoBiNet~\cite{MoBiNet}, we add SI shortcuts with learnable parameters inside CNN blocks to complement block-wise residuals. Our method also performs better than MoBiNet on ImageNet, \emph{e.g.}, 60.45\% \emph{vs.} 54.40\%.

\section{Problem Formulation}~\label{sec:formulation}
Convolutional block is the basis of feature learning in CNN. Each convolutional block in real-value network generally consists of convolutional kernels, activations functions, as well as Batch Normalization (BN)~\cite{ioffe2015BN} layers, \emph{etc}. Given an input image $I$, a feature $O_N$ can be extracted from $I$ by stacking $N$ convolutional blocks. We denote the computation of $O_N$ as,
\begin{eqnarray}
O_N = \mathbf{R}_N(...~\mathbf{R}_{2}(\mathbf{R}_{1}(I))...),
\end{eqnarray}
where $\mathbf{R}_i(\cdot)$ is the $i$-th convolutional block with real-value parameters. $O_N$ can be used for classification~\cite{krizhevsky2012alexnet,VGG,he2016resnet}, segmentation~\cite{long2015fully,Zhu_2019_ICCV,Huang_2019_ICCV}, or detection~\cite{girshick2015fast,He_2017_ICCV}.

Our goal is to simulate the behavior of the real-value network using a binary network with similar structure, which can be denoted as, \emph{i.e.},

\begin{eqnarray}~\label{eq:mainbranch}
\bar O_N = \mathbf{B}_N(...~\mathbf{B}_{2}(\mathbf{B}_{1}(I, \theta_{1}),\theta_{2})...,\theta_N),
\end{eqnarray}
where $\bar O_N$ represents the output of the $N$-th binary convolutional block. $\mathbf{B}_i(\cdot, \theta_i)$ denotes the $i$-th binary convolutional block and $\theta_i$ represents its binary parameters. Note that, because of BN~\cite{ioffe2015BN} layers in $\mathbf{B}_i(\cdot, \theta_i)$, the produced feature $\bar O_N$ can be a real-value tensor.

As discussed in many works~\cite{BNN,rastegari2016xnor,liu2018bireal}, the binary network can be optimized by Back-Propagating (BP) training loss computed with $\bar O_N$ to update each $\mathbf{B}_i(\cdot, \theta_i)$. This training can be achieved by maintaining real-value parameters $\theta_i^*$, updating $\theta_i^*$ with BP loss, and binarizing $\theta_i^*$ to $\theta_i$. Because of quantization errors and vanishing gradients, this training strategy is not efficient in optimizing $\mathbf{B}_i(\cdot, \theta_i)$, especially for convolutional blocks far from the output layer.

To seek a more efficient training strategy, we propose to supervise each binary convolution block with additional cues. Inspired by recent works on distillation learning~\cite{polino2018model,mishra2017apprentice,Zagoruyko2017AT}, we leverage distillation loss derived from FCNN. In other words, each $\mathbf{B}_i(\cdot, \theta_i)$ is enforced to produce similar output with $\mathbf{R}_i(\cdot)$. This block-wise distillation loss trains binary convolution block in a more straightforward way. The distillation loss for the $i$-th convolutional block can be denoted as,
\begin{eqnarray}
\mathcal L_i^{D} = \operatorname D(O_i, \bar O_i),
\end{eqnarray}
where $\operatorname D(\cdot)$ computes distillation loss based on outputs from $\mathbf{R}_i(\cdot)$ and $\mathbf{B}_i(\cdot, \theta_i)$.

Directly optimizing $\mathcal L_i^{D}$ could be difficult because of the limited model capability of binary convolutional kernels. This leads to substantial residuals when comparing $O_i$ and $\bar O_i$. To facilitate the optimization of distillation loss, we further introduce shortcut branches into $\mathbf{B}_i(\cdot, \theta_i)$ to complement residuals of feature maps. Introducing $K$ branches updates the original $\mathbf{B}_i(\cdot, \theta_i)$ as $\bar{\mathbf{B}}_i(\cdot)$, \emph{i.e.},
\begin{eqnarray}
\begin{split}
\label{eqn:binout}
\bar{\mathbf{B}}_i(\bar O_{N-1})=\mathbf{B}_i(\bar{O}_{N-1}, \theta_i)+
\sum_{k=1}^K \mathbf{b}_i(\bar{O}_{N-1}, \gamma_i^{(k)}),
\end{split}
\end{eqnarray}
where $\mathbf{b}_i(\cdot, \gamma_i^{(k)})$ denotes the $k$-th shortcut branch with binary parameters $\gamma_i^{(k)}$.

As shown in Eq.~\eqref{eqn:binout}, $K$ branches are used to model the residuals between $\mathbf{R}_i(\cdot)$ and $\mathbf{B}_i(\cdot, \theta_i)$. With a properly trained $\mathbf{B}_i(\cdot, \theta_i)$, the residual would exhibit limited complexity and variation. This makes it possible to compress the $K$ branches and limit their parameter overheads below a given threshold $\epsilon$. We hence represent the overall training loss as,

\begin{eqnarray}
\begin{split}
\min \limits_{\Theta}  \mathcal{L}= &\mathcal L^{BP} + \alpha\sum_{i=1:N}\mathcal{L}_i^D,
\text{     subject to }\sum_{i=1:N}\sum_{k=1:K}|\gamma_i^{(k)}|<\epsilon,
\end{split}
\end{eqnarray}
where $\Theta$ denotes the collection of parameters in binary network and $\mathcal L^{BP}$ denotes the loss at output layer. $\epsilon$ denotes the limitation to memory overheads introduced by shortcut branches. $ \alpha $ denotes the loss weight.

\section{Implementation}
~\label{sec:method}
This section first presents our main branch structure, then proceeds to introduce the Squeeze-and-Interaction (SI) shortcut and block-wise distillation loss.


\begin{figure}[t]
\centering
		\includegraphics[width=4.8in]{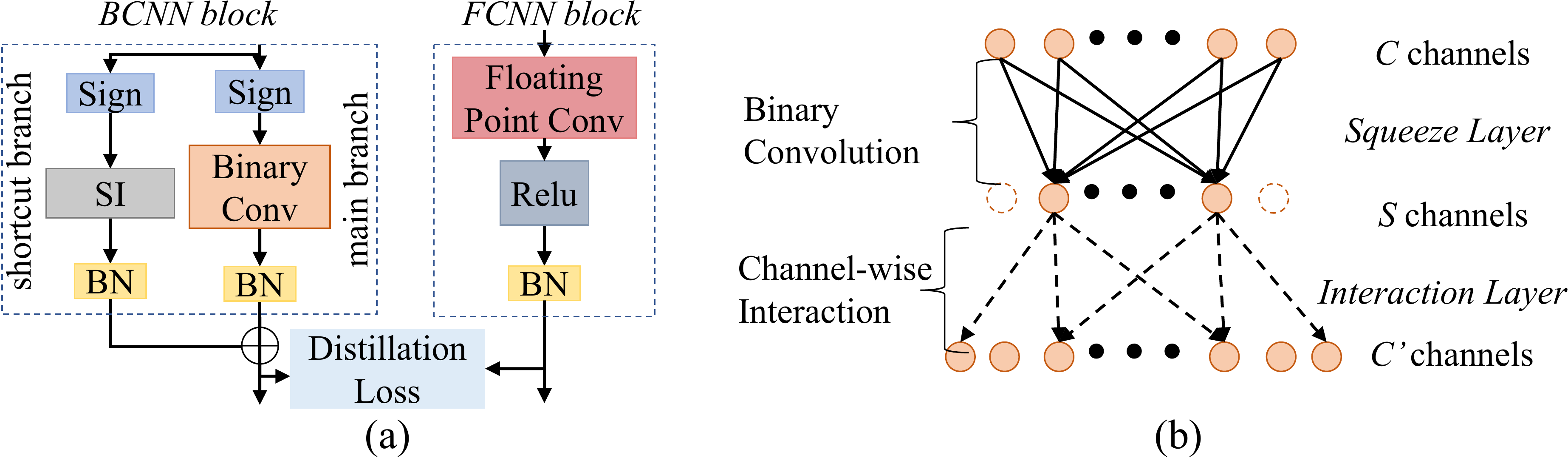}
		\caption{Illustration of our convolutional block in (a) and Squeeze-and-Interaction (SI) shortcut in (b). Our convolution block consists of a main branch and $K$($K=1$) SI shortcut branches. SI computes a $S$-channel feature map with the squeeze layer, then generates a $C'$-channel output with the interaction layer, $S<C'$. }
	\label{fig:net}
\end{figure}
\subsection{Structure of Main Branch}
The main branch implements the computations in Eq.~\eqref{eq:mainbranch}. Fig.~\ref{fig:net}(a) illustrates the structure of a binary convolutional block in the main branch. It contains a sign function to convert a real-value tensor into a binary one, which is hence computed with a binary convolution kernel. The output is input into a Batch Normalization layer to produce a real-value tensor as the output.


\textbf{Biased sign function}: Sign function converts real-value tensors to binary ones and causes considerable quantization errors. Existing networks~\cite{BNN,liu2018bireal,rastegari2016xnor} implement the sign function with a fixed threshold $0$, \emph{i.e.}, values larger than $0$ are quantized to 1, otherwise -1. In order to decrease the quantization errors, we introduce a trainable threshold $t$ to implement a biased sign function into each convolutional block. This biased sign function can be denoted as,
\begin{eqnarray}
\begin{aligned}
\label{eq:sign}
\operatorname{sign}(x, t) =
\left\{
\begin{matrix}
-1&\text{ if  $x \leq t$},\\
1&\text{ if  $x > t$},
\end{matrix}\right.
\end{aligned}
\end{eqnarray}
where $t$ is the trainable threshold. The effects of $t$ will be discussed in Sec.~\ref{sec:abstudy}.

\textbf{Forward propagation}: We follow the method in previous works~\cite{BNN} to implement the forward prorogation of main branch. For the $i$-th convolutional block, the main branch first binarizes the input tensor $\bar{O}_{i-1}$ with the sign function, Then, the binary convolution is computed with XNOR and bitcount operations. We represent the forward propagation of main branch in a binary convolutional block as,
\begin{eqnarray} \label{eq:forward}
\bar{O}_i=\operatorname {BN}(\operatorname{sign}(\bar{O}_{i-1}, t)\otimes \theta_i)), \theta_i = \operatorname{sign}(\theta_i^*, 0),
\end{eqnarray}
where $\otimes$ denotes convolution, $\theta_i^*$ is the floating point parameters used for loss back propagation and parameter updating. sign function converts $\theta_i^*$ into binary convolutional kernel $\theta_i$. $\operatorname {BN}(\cdot)$ is the Batch Normalization.

Compared with the convolutional block in FCNN, the one in BCNN omits the ReLU layer and accelerates the computation with binary convolutions. Our training stage learns proper $t$, $\theta_i^*$, and BN parameters, to simulate the convolutional block in FCNN. We proceed to introduce the SI shortcut and block-wise distillation loss to facilitate the optimization.

\subsection{Squeeze-and-Interaction Shortcut Branch} \label{sec:shortcut}

As shown in Fig.~\ref{fig:net}(a), a Squeeze-and-Interaction (SI) shortcut branch is trained to produce a residual feature map, which is hence fused with the feature map from main branch. Computed based on a properly trained main branch, the target residual feature map would exhibit limited variances. This makes it possible to model residuals with a lightweight shortcut branch.

This intuition leads to the structure in Fig.~\ref{fig:net}(b). We take VGG-small network as an example. For an input tensor $\bar O_{i-1}$ with $C$ channels, the shortcut branch first converts it into a binary tensor with Eq.~\eqref{eq:sign}. A squeeze layer uses $S \times 3\times 3 \times C$ sized binary convolutional kernel to produce a $S$ ($S<C'$) channel feature map. Then, this feature map is mapped to a $C'$ channel feature map by interaction layer with sparse channel-wise connections.

\textbf{Squeeze Layer}: The squeeze layer is learned by firstly predicting a $C'$ channel feature map with a $C' \times 3\times 3 \times C$ sized binary convolutional kernel, then keeping $S$ channels and discarding the others. There are many ways to select $S$ channels, \emph{e.g.}, through random selection. We perform automatic selection by introducing a learnable $C'$-dim real-value weighting vector $\omega$. $\omega$ is learned to weight the importance of each channel in an end to end training. The computation of a squeeze layer can be represented as,
\begin{eqnarray}
\label{eqn:res_squeeze}
O_{sqz}[c]=\operatorname {BConv}({O}_{in})[c]\times \omega[c], c=1:C'
\end{eqnarray}
where $\operatorname {BConv}(\cdot)$ denotes the binary convolution with kernel size $C' \times 3\times 3 \times C$. During end to end training, $\omega$ could be learned together with the $\operatorname {BConv}$. The weight vector $\omega$ provides cues about importance of each channel, $e.g.$, a channel $c$ would be more important, if $|\omega [c]|$ is larger. We hence could select and keep important channels according to the absolute values in $\omega$.

Suppose we introduce $K$ shortcut branches into each block, to keep the overall memory overheads ration below a given threshold $\epsilon \in[0,1]$ compared to the main branch. We need to select
$n = \epsilon \times \sum_{i=1:N}C_i'$ channels to keep and discard the others. This can be conducted by a block-wise selection, which selects top $C_i'\times \epsilon$ channels with largest weights from the $i$-th block. A global selection strategy can also be conducted by selecting the top $\epsilon \times \sum_{i=1:N}C_i'$ channels in the BCNN. Different selection strategies will be tested in Sec.~\ref{sec:abstudy}.

\textbf{Interaction Layer}:
After selecting $S$ channels, the interaction layer introduces a channel-wise sparse interaction to generate a feature map with $C'$ channels. This is achieved by learning a real-value sparse matrix $T$ with size $S\times C'$. The computation of interaction layer can presented as,
\begin{eqnarray}
\label{eqn:res_inter}
O_{itr}=O_{sqz}\times T,
\end{eqnarray}
where $O_{itr}$ is the $C'$ channel output by interaction layer.

To ensure high computation efficiency in interaction layer, $T$ is kept sparse. It is initialized as a sparse matrix with $S$ non-zero elements, where $T[i][j]$ = 1 only if $i$ and $j$ correspond to the same channel. After end-to-end training, small values in $T$ are set to 0 to ensure high sparsity. After selecting $S$ channels in Squeeze layer and fixing $T$, shortcut branch is fine-tuned to recover performance.

\textbf{Discussions}:
As discussed above, $S$ at each shortcut branch is determined by our channel selection with $\omega$. This strategy selects different channels for different CNN blocks, \emph{e.g.}, more channels are kept for important blocks. Another way is directly training the shortcut branches with a given $S$. Compared with the given $S$, our strategy has potential to achieve better performance with the same memory overheads.
More discussions will be presented in Sec.~\ref{sec:abstudy}.

\subsection{Block-wise Distillation Loss}

\begin{figure}[t]
	\centering
	\includegraphics[width=3.3in]{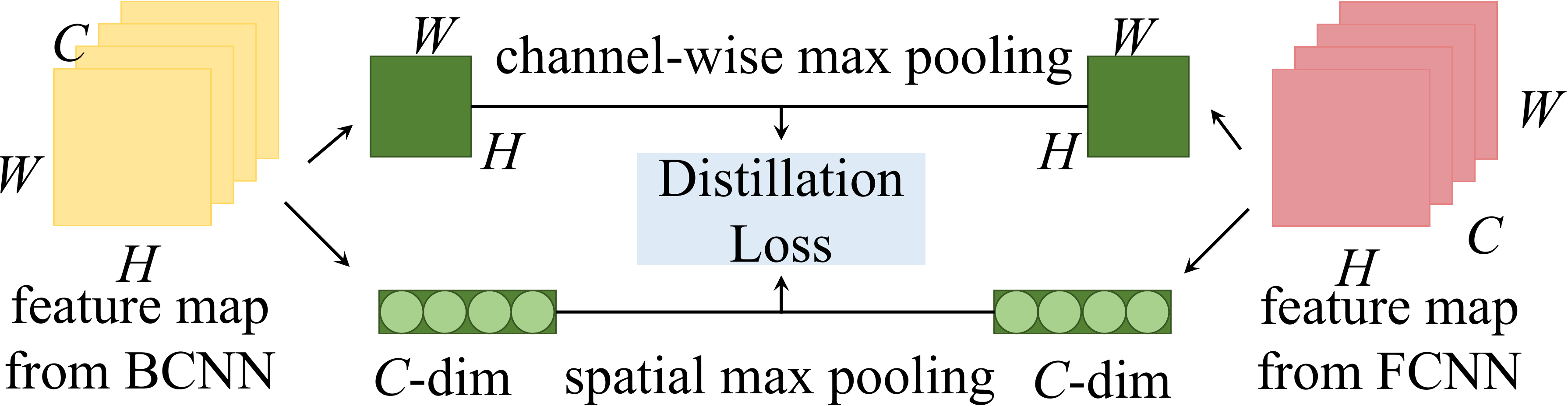}
	
	\caption{Illustration of our block-wise distillation loss, which is computed with channel-wise and spatial-wise max pooling.}
\vspace{-3mm}
	\label{fig:distillation}
\end{figure}
As shown in Fig.~\ref{fig:net}(a), to facilitate the optimization of BCNN block, we compute distillation loss by referring to both the spatial and channel differences. This differs with many existing methods that train CNN by computing distillation loss at the end of the network~\cite{polino2018model,mishra2017apprentice}. Some other works~\cite{Zagoruyko2017AT} compute the distillation loss on feature maps after spatial pooling. Only considering the spatial pooling fails to mine the channel-wise differences.

As shown in Fig.~\ref{fig:distillation}, for $W\times H \times C$ sized feature maps, we first compute $W\times H$ sized spatial maps and $C$-dim vectors with spatial and channel-wise pooling. Then, the distillation loss on the $i$-th convolutional block can be computed between BCNN and FCNN. The computation can be denoted as,
\begin{eqnarray}
\begin{split}
{\mathcal L^{D}_i}&=||\frac{\operatorname{SP}(O_{i})}{||\operatorname{SP}(O_{i})||_2} - \frac{\operatorname{SP}(\bar{O}_{i})}{||\operatorname{SP}(\bar{O}_{i})||_2}||_2
+||\frac{\operatorname{CP}(O_{i})}{||\operatorname{CP}(O_{i})||_2} -  \frac{\operatorname{CP}(\bar{O}_{i})}{||\operatorname{CP}(\bar{O}_{i})||_2}||_2,
\end{split}
\end{eqnarray}
where $\operatorname{SP}(\cdot)$ denotes spatial-wise max pooling and $\operatorname{CP}(\cdot)$ denotes channel-wise max pooling. $||\cdot||_2$ computes the $L_2$ norm of a vector.

Compare to method~\cite{Zagoruyko2017AT}, the proposed distillation loss considers extra channel-wise difference. Another work~\cite{zhuang2018towards} computes distillation loss on the entire feature map by measuring the element-wise difference. Compared with ~\cite{zhuang2018towards}, our method could be more robust by involving pooling strategies. Different distillation loss functions are tested in Sec.~\ref{sec:abstudy}.


\subsection{Training and Optimization}
\textbf{Overall Loss Function}: For each convolution block in BCNN, the loss is composed by a block-wise distillation loss and the task specific loss back-propagated by its subsequent layers. Therefore, our method is general and shows potentials to train BCNNs for different vision tasks. We conduct experiments on image classification task. The overall loss function can be formulated as

\begin{eqnarray}\label{eqn:loss}
{\mathcal L}={\mathcal L^{BP}}(\bar O_N, y) + \alpha\sum_{i=1:N}{\mathcal L^{D}_i}(O_{i}, \bar O_{i}),
\end{eqnarray}
where $y$ is the ground-truth image-label and $\bar O_N$ is feature for classification. $L^{BP}(\cdot)$ computes the image classification loss at output layer. We use cross-entropy loss to implement $L^{BP}(\cdot)$. Loss weight $ \alpha $ is set as $0.1$ referring to~\cite{Zagoruyko2017AT}.

\textbf{Optimization}: In forward-propagation shown in Eq.~\eqref{eq:forward}, the $\operatorname{sign}(x)$ function is not derivable at $x=0$. We refer to the straight-through estimator~\cite{BNN} for network training. We use real-value parameters $\theta^*$ for back-propagation and network training. For the $i$-th convolutional block, derivatives of loss with respective to network parameters is computed as,
\begin{eqnarray}
\begin{split}
\frac{\partial {\mathcal L}}{\partial \theta_{i}^*}=&
\alpha\frac{\partial {\mathcal L^{D}_i}}{\partial \theta_{i}^*}+ \frac{\partial {\mathcal L^{BP}}}{\partial \theta_{i}^*}
=(\alpha\frac{\partial {\mathcal L^{D}_i}}{\partial \bar{O}_{i}}  +\frac{\partial L^{BP}}{\partial \bar{O}_{i}})\frac{\partial \bar{O}_i}{\partial \operatorname{sign}(\theta_i^*, 0)} \frac{\partial \operatorname{sign}(\theta_i^*, 0)}{\partial \theta_i^*}\\
=&\frac{\partial (\alpha\mathcal L^{D}_i+\mathcal L^{BP})}{\partial \bar{O}_{i}}\frac{\partial \bar{O}_i}{\partial \operatorname{sign}(\theta_i^*, 0)} 1_{|\theta_i^*|<1}.
\end{split}
\end{eqnarray}

For back-propogation, derivative of input of previous block is calculated as:
\begin{eqnarray}
\begin{split}
\frac{\partial (\alpha\mathcal L^{D}_i+\mathcal L^{BP})}{\partial \bar{O}_{i}} &=\frac{\partial (\alpha\mathcal L^{D}_i+\mathcal L^{BP})}{\partial \operatorname{sign}(\bar{O}_{i}, b)} \frac{\partial \operatorname{sign}(\bar{O}_{i}, b)}{\partial \bar{O}_{i}}=\frac{\partial (\alpha\mathcal L^{D}_i+\mathcal L^{BP})}{\partial \operatorname{sign}(\bar{O}_{i}, b)} 1_{|\bar{O}_{i}-b|<1}.
\end{split}
\end{eqnarray}

Since $\bar{O}_{i}$ is the sum of the outputs from main branch and SI shortcut branches. The back-propagation functions of ${\partial {\mathcal L^{D}_i}}/{\partial \gamma_i^{(k)}}$ and ${\partial {\mathcal L^{BP}}}/{\partial \gamma_i^{(k)}}$, $k\in 1,2...K$ share a similar formulation as  ${\partial {\mathcal L^{D}_i}}/{\partial \theta_i}$ and ${\partial {\mathcal L^{BP}}}/{\partial \theta_i}$ respectively.

\section{Experiments}

\subsection{Dataset}
We conduct experiments on two widely used datasets to evaluate BCNN performance: CIFAR-10~\cite{cifar10} and ImageNet~\cite{deng2009imagenet}. CIFAR-10 consists of 60k $32\times 32$ sized images in 10 classes, including 50k training images and 10k test images. ImageNet is a large-scale dataset with 1,000 classes and 1.2 million training images, as well as 50k validation images. We use DGRL to represent our BCNN.
\subsection{Implementation Details}
We trains two BCNNs according to VGG-small and ResNet18, respectively. Note that, existing works~\cite{liu2018bireal,wang2019CI-BCNN} do not binarize the first conv layer, the last fc layer, as well as $ 1\times 1 $ convolutions in ResNet. We also follow this setting. The following parts present details for training of those two BCNN.

\textbf{VGG-small}: We conduct all experiments on CIFAR-10 using VGG-small~\cite{BNN}.
We pad 4 pixels on each side of the input image and crop it randomly into the size of $ 32 \times 32$. Then, all images are scaled into the range $[-1, 1]$. We use the block structure as in XNOR-Net~\cite{rastegari2016xnor} as the main branch. The real-value VGG-small network proposed in~\cite{VGG} is used as the FCNN for BNN training. FCNN is trained on CIFAR-10. We regard each convolutional layer with a ReLU layer and BN layer as a convolutional block. This leads to 5 convolutional blocks and 5 block-wise distillation loss computed. We train BCNN from scratch and follow the settings in XNOR-Net~\cite{rastegari2016xnor}. We set batch size as $128$ and initial learning rate as $0.01$. We train it for $320$ epochs and reduce the learning rate by $ 0.1 $ at epoch $120$, $200$, $240$ and $280$, respectively.

\textbf{ResNet18}: On ImageNet, we build BCNN with ResNet18. For each input image, a $224 \times 224$ region is randomly cropped for training from the resized image whose short side is $256$. We use the basic block in BinaryResNetE~\cite{bethge2019Simplicity} to implement our main branch. The ResNet proposed in~\cite{he2016resnet} is used as the corresponding FCNN, which is trained on ImageNet. ResNet18 has 4 convolutional blocks. Our binary ResNet18 hence has 4 binary convolutional block and is trained with 4 block-wise distillation loss. During training, we set the batch size as $1024$ and the initial learning rate as $0.004$. We train the BCNN from scratch for $120$ epochs and reduce the learning rate by $ 0.1 $ at epoch 70, 90 and 110, respectively.

\begin{figure}[t]
	\centering \includegraphics[width=0.9\linewidth]{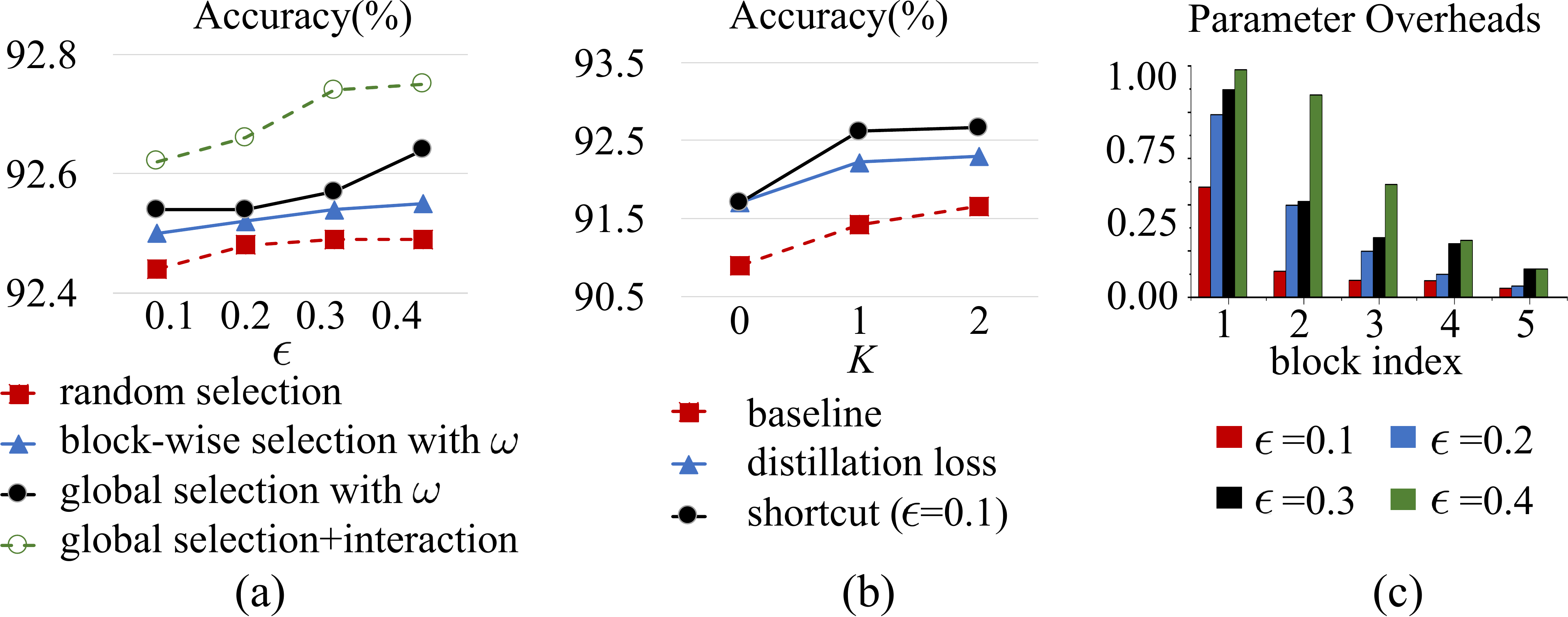}\label{fig:shortcut}
	\caption{Validity of channel selection in (a) and branch number $K$ in (b). (c) shows parameter overheads in each block after global channel selection.}\label{fig:shortcut}
\end{figure}

\subsection{Ablation Study}
\label{sec:abstudy}
This part first tests the validity of SI shortcuts, then analyzes the choosing of branch number $K$, as well as our training strategies. Experiments are conducted on CIFAR-10 with VGG-small.


\textbf{Channel Selection in SI shortcut}: Channel number $S$ can be selected with different strategies, \emph{e.g.}, random selection, block-wise selection with $\omega$, and global selection with $\omega$. This part tests those strategies with different parameter overheads, \emph{i.e.}, $\epsilon$ and presents the comparison in Fig.~\ref{fig:shortcut}(a). It is clear that, random selection gets the worst performance. This implies that, different channels present varied importance, thus should not be randomly selected. Layer-wise selection with $\omega$ outperforms the random selection, indicating the validity of weight vector $\omega$. Global selection further outperforms block-wise selection. This shows that different CNN blocks present different importance. To maintain the same parameter overheads, more channels should be kept for important blocks.

Fig.~\ref{fig:shortcut}(a) shows that, larger $\epsilon$ does not bring substantial performance boost over $\epsilon = 0.1$, indicating that the block-wise residuals can be effectively modeled with a lightweight shortcut. Fig.~\ref{fig:shortcut}(a) also shows the performance after inserting the Interaction layer, which brings more substantial performance gains than larger $\epsilon$, showing the importance of Interaction layer. According to Fig.~\ref{fig:shortcut}(a), we fix $\epsilon = 0.1$ to ensure network compactness.

\begin{table} [t]
	\centering
	\caption{Validity of block-wise distillation loss $\mathcal L^D$, biased sign function, and step training strategy. $K=1$ means adding a shortcut branch with $\epsilon =0.1$.}\label{tab:abstudy}
	\begin{center}
		\setlength{\tabcolsep}{3mm}{\resizebox{0.9\textwidth}{!}{
			\begin{tabular}{c|cccccc}
				\hline
				distillation&- & spatial & $\mathcal L^D$ & $\mathcal L^D$ & $\mathcal L^D$ & $\mathcal L^D$\\\hline
				\textit{K}&0&0&0&1&1&1\\\hline
				sign func.&$t$=0&$t$=0&$t$=0&$t$=0&learned $t$&learned $t$\\\hline
				step train&-&-&-&-&-&\checkmark\\\hline
				accuracy(\%)&90.90&91.32&91.71&92.25&92.41&\textbf{92.62}\\
				\hline
			\end{tabular}
		}}
	\end{center}
\vspace{-4mm}
\end{table}

\textbf{Shortcut Branch Number $K$}: Each convolutional block allows to introduce $K$ shortcut branches. Introducing more branches potentially helps to complement the block-wise residuals with FCNN. We hence test the parameter $K$ and present the results in Fig.~\ref{fig:shortcut}(b). We use XNOR-Net~\cite{rastegari2016xnor} as the baseline, and first insert shortcut branches having the same structure with main branch. Therefore, introducing 2 shortcut branches leads to 200\% parameter overheads. As shown in Fig.~\ref{fig:shortcut}(b), adding more branches to baseline boosts its performance. Meanwhile, block-wise distillation loss is also beneficial for performance boost. Fig.~\ref{fig:shortcut}(b) also compares the performance of inserting SI branches which involve channel selection with parameter overhead $\epsilon = 0.1$. It can be observed, SI branch effectively drops the memory overheads, meanwhile boosts the performance. Also, channel selection with a fixed $\epsilon$ is not sensitive to a larger $K$. We hence fix $K=1$ in following experiments.

\textbf{Training Strategies}: Table~\ref{tab:abstudy} further shows the validity of distillation loss, biased sign function, and step training strategy, which are valid in boosting BCNN performance. Distillation loss computed with spatial and channel pooling, \emph{i.e.}, $\mathcal L^D$, outperforms the one computed only with spatial pooling. $\mathcal L^D$ brings substantial performance gains over the baseline. Biased sign function with learned $t$ outperforms the one with fixed $t=0$. Our step training, \emph{i.e.}, first train and fix main branch then train shortcut branch, further brings performance gains. Combination of those strategies boosts the accuracy from 90.9\% to 92.62\%. Note that, in each block we use addition to fuse features from main and shortcut branches. This simple fusion strategy fits well for our step training strategy.

\begin{figure}[t]
	\centering
	\includegraphics[width=1\linewidth]{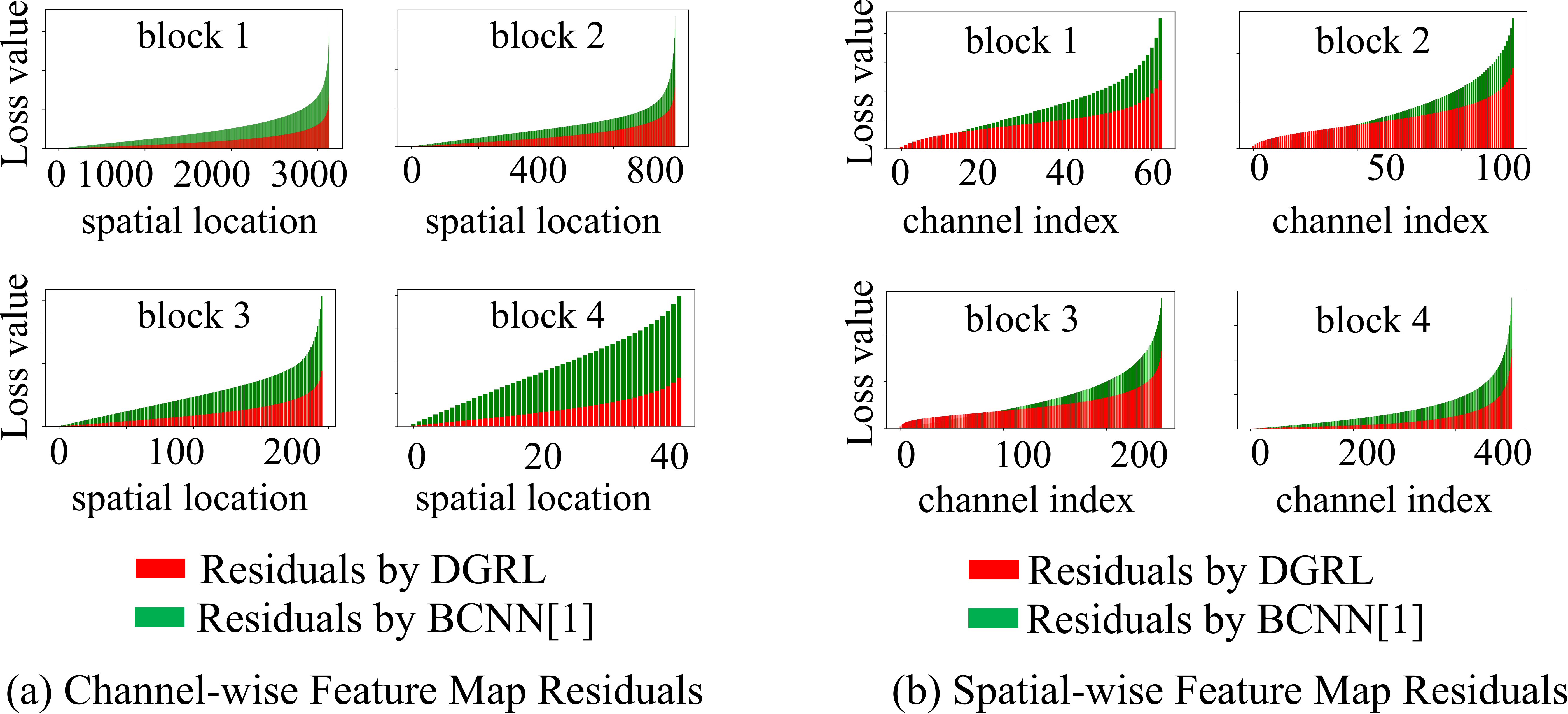}
	\caption{Statistics of feature map residuals on ImageNet. Different blocks have different spatial sizes and channel numbers.}	
	~\label{fig:statistics}
\vspace{-2mm}
\end{figure}

\textbf{Discussions}: Fig.~\ref{fig:shortcut}(c) analyzes parameter overheads introduced to each block by the global channel selection. With different $\epsilon$, our method tends to keep more channels for shallow CNN blocks. This could be because shallow blocks are the basis for learning discriminative features. Well-trained shallow blocks also help to alleviate the accumulated residuals as network goes deeper. Fig.~\ref{fig:statistics} shows statistics of feature map residuals on ImageNet, where both spatial and channel-wise residuals between BCNN and FCNN are illustrated. It is clear that, our BCNN, \emph{i.e.}, DRGL, produces smaller residuals than baseline BCNN~\cite{bethge2019Simplicity}, indicating our BCNN could better simulate the responses in FCNN.

\subsection{Comparison with Recent Works}


~\label{sec:sota}
\textbf{Comparison on CIFAR-10:}
We compare with quantized networks including BWN~\cite{rastegari2016xnor}, TWN~\cite{li2016TWN}, TBN~\cite{Wan2018TBN}, and BCNN including BNN~\cite{BNN}, XNOR-Net~\cite{rastegari2016xnor}, CI-BCNN~\cite{wang2019CI-BCNN}. The comparisons are summarized in Table~\ref{tab:sota} (a).

Table~\ref{tab:sota} (a) shows that our method achieves promising performance compared with quantized networks, which present higher computation and memory complexities. Only using main branch with $K$=0, our DGRL outperforms BWN~\cite{rastegari2016xnor} and TBN~\cite{Wan2018TBN}, and achieves comparable performance with TWN~\cite{li2016TWN}, \emph{e.g.}, our 92.29\% \emph{vs.} 92.56\% of TWN with about 2\% FLOPs of TWN. With a shortcut branch, DGRL achieves accuracy of 92.62\%, which outperforms TWN~\cite{li2016TWN}.

DGRL also outperforms other binary networks in Table~\ref{tab:sota} (a). DGRL with $K$=0 share the same structure with baseline XNOR-Net~\cite{rastegari2016xnor}. Our training strategy makes DRGL outperforms the baseline by $2.39\%$, showing the effectiveness of our distillation loss. We also show the performance using the SI shortcut without distillation loss, which also outperforms XNOR-Net~\cite{rastegari2016xnor}. DGRL with $K$=1 and distillation loss further outperforms the recent CI-BCNN on CIFAR-10 with similar computational and memory complexities.
\begin{table} [t]
	\caption{Comparison of with recent works. W, A denote the precision of network parameters and activations, respectively.}~\label{tab:sota}
	
	\begin{minipage}{1\linewidth}
		\vspace{2mm}
		\begin{center}
			{\small {(a) Comparison on CIFAR-10 with VGG-small.}}
			
			\setlength{\tabcolsep}{3mm}{
				\vspace{1mm}
				\resizebox{1\textwidth}{!}{
					
				\begin{tabular}{c|c|cc|c|c}
					\hline
					Methods	&accuracy (\%)	&W (bits)	&A (bits)	&FLOPs& Size (Mbits)\\ \hline
					Full-precision	&93.20&32	&32	&  $6.17\times 10^8$ &428.96\\
					\hline
					BWN~\cite{rastegari2016xnor}	&90.10&1	&32	&  $3.09\times 10^8$ & 14.80\\
					TWN~\cite{li2016TWN}	&92.56&2	&32	&$6.17\times 10^8$ &28.16\\
					TBN~\cite{Wan2018TBN}	&90.87&1	&2	 & $1.80\times 10^7$ &14.80\\
					\hline
					BNN~\cite{BNN}	&89.90&1	&1	&$1.32\times 10^7$  &14.80 \\
					XNOR-Net~\cite{rastegari2016xnor}	&89.80&1	&1	&$1.32\times 10^7$ &14.80\\
					CI-BCNN~\cite{wang2019CI-BCNN}	&92.47&1	&1	&$1.32\times 10^7$  &14.80\\
					DGRL ($K$=$0$) & 92.29 & 1&1 & ${1.32}\times {10}^{7}$ & 14.80\\
					DGRL ($K$=$1$) w/o ${\mathcal L^{D}}$& {91.59}&{1} &{1} & {$1.48\times 10^7$ }& {14.84}\\
					\textbf{DGRL ($K$=$1$)}& \textbf{92.62}&{1} &{1} & {$1.48\times 10^7$ }& {14.84}\\
					\hline
				\end{tabular}
			}}
		\end{center}
	\end{minipage}

	\begin{minipage}{1\linewidth}
		\vspace{3mm}
		\begin{center}
			\small {(b) Comparison on ImageNet with ResNet18. $^\dagger $ denotes further compressing $1\times1$ convolutions in SI branch with channel selection. }
			
			\setlength{\tabcolsep}{3mm}{
				\vspace{1mm}
			\resizebox{1\textwidth}{!}{
				
				\begin{tabular}{c|c|cc|c|c}
					\hline
					Methods &accuracy (\%)&	W (bits)&	A (bits)  &  FLOPs &   Size (Mbits) \\\hline
					Full-precision  &69.30& 32 & 32 & $1.81 \times 10^9$& 374.1\\ \hline
					BWN~\cite{rastegari2016xnor} &60.80&1&32&$9.20 \times 10^8$ & 33.7\\
					DoReFa-Net~\cite{zhou2016dorefa} &59.20 & 1& 4&$4.91\times10^8$&182.72\\
					ABC-Net~\cite{lin2017ABC-Net}&49.10&3&1&$2.53\times 10^8$&81.6\\
					BENN~\cite{zhu2019BENN}&53.60&1&1&$5.01 \times 10^8$&101.1\\
					\hline
					XNOR-Net~\cite{rastegari2016xnor}&51.20&1&1&$1.67 \times 10^8$&33.7\\
					ABC-Net~\cite{lin2017ABC-Net}&42.70&1&1&$1.05\times 10^8$&27.2\\
					Bi-Real-Net~\cite{liu2018bireal}&56.40&1&1&$1.63 \times 10^8$&33.6\\
					PCNN~\cite{gu2019projection}&57.30&1&1&$1.67 \times 10^8$&33.7\\
					RBCN~\cite{liu2019ganbin}&59.50 & 1&1& $1.67 \times 10^8$ &33.7\\
					BONNs~\cite{gu2019bayesian} &59.30 &1&1& $1.63 \times 10^8$ &33.6 \\
					BinaryResNetE18~\cite{bethge2019Simplicity} & 58.10 & 1&1&$1.63 \times 10^8$&33.6 \\
					CI-BCNN~\cite{wang2019CI-BCNN}&59.90&1&1&$1.63 \times 10^8$&33.6	\\
					MoBiNet-Mid~\cite{MoBiNet} & 54.40 & 1 &1 & $0.52 \times 10^8$&25.1 \\
					\hline
					DGRL ($K$=$0$)	&59.50&1&1&${1.63} \times  {10}^{8}$&{33.6}\\
					DGRL($K$=$1$) w/o ${\mathcal L^{D}}$ & 59.83&1&1 &${1.84} \times 10^8$ & 35.5\\
					\textbf{DGRL ($K$=$1$)}&\textbf{60.45}&{1}&{1}&	${1.84} \times 10^8$ & 35.5	\\
					\textbf{DGRL$^\dagger $ ($K$=$1$)}&\textbf{60.23}&{1}&{1}&	${1.68} \times 10^8$ & 34.6	\\\hline
			\end{tabular}}}
		\end{center}
	\end{minipage}
\end{table}

\textbf{Comparison on ImageNet}: We compare DGRL with quantized neural networks including BWN~\cite{rastegari2016xnor}, DoReFa-Net~\cite{zhou2016dorefa}, ABC-Net~\cite{lin2017ABC-Net} with 3 bases, and binary neural networks including XNOR-Net~\cite{rastegari2016xnor}, ABC-Net~\cite{lin2017ABC-Net} with 1 base, Bi-Real-Net~\cite{liu2018bireal}, BENN~\cite{zhu2019BENN}, PCNN~\cite{gu2019projection}, RBCN~\cite{liu2019ganbin}, BONNs~\cite{gu2019bayesian}, BinaryResNetE18~\cite{bethge2019Simplicity}, CI-BCNN~\cite{wang2019CI-BCNN}, MoBiNet-Mid~\cite{MoBiNet}. Table~\ref{tab:sota} (b) shows the comparisons.

Comparisons with quantized networks in Table~\ref{tab:sota} (b) shows similar conclusion with that in Table~\ref{tab:sota} (a). DGRL ($ K$=$1$) achieves comparable performance with BWN~\cite{rastegari2016xnor} using 20\% of its FLOPs. DGRL outperforms the other three quantized networks in aspects of classification accuracy, FLOPs, and model size. For example, ABC-Net~\cite{lin2017ABC-Net} involves 3 convolutional bases. BENN~\cite{zhu2019BENN} ensembles outputs from 3 BCNNs. DRGL features more efficient design and better performance than those networks.

DGRL ($K$=1) consistently outperforms the other binary networks. DGRL with $K$=0 share identical structure with BinaryResNetE18~\cite{bethge2019Simplicity}. Our training strategy, \emph{e.g.}, block-wise distillation loss boosts the baseline performance by $1.40\%$. Adding one SI shortcut boosts the performance by 1.73\%. Combining SI shortcut and distillation loss achieves the performance of 60.45\%, outperforming the state-of-the art CI-BCNN~\cite{wang2019CI-BCNN} by 0.55\%. RBCN~\cite{liu2019ganbin} trains the BCNN with a FCNN by training GANs. Our method substantially outperforms RBCN, \emph{i.e.}, 60.45\% \emph{vs.} 59.5\%. Our training strategy is also more straight-forward and efficient than training GANs in RBCN.

\begin{table} [t]
	\caption{FLOPs (in million) in main and SI shortcut branches in DGRL$^ \dagger $ ($K$=1) implemented with ResNet18. FLOPs is computed following~\cite{liu2018bireal}.}~\label{tab:memory}
	\begin{center}\setlength{\tabcolsep}{2.5mm}{
		\resizebox{1\textwidth}{!}{
			\begin{tabular}{c|c|c|c|c|c|c|c|c}
				\hline
				&operation&conv0&block0&block1&block2&block3&fc&sum\\\hline
				\multirow{2}*{main branch}&{binary} & -&7.2& 6.3& 6.3& 6.3&-&26.1\\
				&{float} & 118 & 0.0002 & 6.4 & 6.4 & 6.4 &0.5 &137\\
				\hline
				\multirow{2}*{SI shortcut}&{binary} & - & 0.76 & 0.38&0.049&  0.001 & -&1.2 \\
				&{float} & - &0.0083& 2.71 &0.35 & 0.01& - & 3.1\\
				\hline
		\end{tabular}}}
	\end{center}
\end{table}

\textbf{Efficiency and Memory Usage}: Efficiency and memory usage are important for BCNNs. Table~\ref{tab:sota} compares the FLOPs and model size across different networks. The model size and FLOPs are computed following~\cite{liu2018bireal}, where FLOPs is computed as the amount of floating point multiplications plus 1/64 of the amount of 1-bit multiplication. Table~\ref{tab:sota} shows that DGRL achieves competitive performance with reasonably good computation and memory efficiency. Compared with full-precision network on CIFAR-10, our method achieves slightly lower accuracy, \emph{i.e.}, 92.62\% \emph{vs.} 93.20\%, but significantly saves FLOPs by about $40\times$, and saves memory by about $29\times$. Compared with quantized networks on ImageNet, DGRL shows substantial advantages in efficiency and compactness.
It also outperforms other BCNN with comparable efficiency and compactness.

Table~\ref{tab:sota} shows that, the SI shortcut branch introduces marginal computational and parameter overheads, \emph{e.g.}, 5.7\% and 12.9\% in model size and FLOPs, respectively over the baseline in Table~\ref{tab:sota} (b). For our binary ResNet18, we can compress the $1 \times 1$ convolutions in SI shortcuts by discarding channels with small weights learned by Eq.~\eqref{eqn:res_squeeze}. This operation further decreases the computations and memory overheads, while maintaining a similar performance, \textit{i.e.}, 60.23\%.

To verify the effects of SI branches to BCNN inference speed, we test BCNNs with/without shortcut branches, and get similar average time to process one image: 2.063ms \emph{vs.} 1.875ms on a 1080TI GPU using cuDNN 7.6.1 with batch size of 16. Thus, shortcut branch does not significantly slow down the speed. This is because main and shortcut branches are computed in parallel. The speed is decided by the slower one. Table~\ref{tab:memory} compares number of binary and floating operations in main and shortcut branches. It is clear that, shortcut branch needs less computations, thus is faster. Table~\ref{tab:memory} also shows that, the speed bottleneck of BCNN is the conv0, which is also not binarized in existing BCNNs~\cite{liu2018bireal,wang2019CI-BCNN}.

\section{Conclusion}
This paper targets to learn a compact BCNN with guidance from FCNN. This is achieved by optimizing each binary convolutional block with block-wise distillation loss derived from FCNN, as well as updating binary convolutional block architecture. The block-wise distillation loss leads to a more effective optimization to BCNN. A lightweight shortcut branch with SI structure is inserted into each binary convolutional block to complement residuals at each block. Extensive experiments on CIFAR-10 and ImageNet demonstrate the superior performance of the proposed method in aspects of both classification accuracy and efficiency.

\clearpage
%
%
\bibliographystyle{splncs04}
\bibliography{egbib}

\begin{thebibliography}{10}
\providecommand{\url}[1]{\texttt{#1}}
\providecommand{\urlprefix}{URL }
\providecommand{\doi}[1]{https://doi.org/#1}

\bibitem{bethge2019Simplicity}
Bethge, J., Yang, H., Bornstein, M., Meinel, C.: Back to simplicity: How to
  train accurate bnns from scratch? arXiv preprint arXiv:1906.08637  (2019)

\bibitem{deng2009imagenet}
Deng, J., Dong, W., Socher, R., Li, L.J., Li, K., Fei-Fei, L.: Imagenet: A
  large-scale hierarchical image database. In: CVPR (2009)

\bibitem{freund1997decision}
Freund, Y., Schapire, R.E.: A decision-theoretic generalization of on-line
  learning and an application to boosting. Journal of computer and system
  sciences  \textbf{55}(1),  119--139 (1997)

\bibitem{girshick2015fast}
Girshick, R.: Fast r-cnn. In: ICCV (2015)

\bibitem{gu2019projection}
Gu, J., Li, C., Zhang, B., Han, J., Cao, X., Liu, J., Doermann, D.: Projection
  convolutional neural networks for 1-bit cnns via discrete back propagation.
  In: AAAI (2019)

\bibitem{gu2019bayesian}
Gu, J., Zhao, J., Jiang, X., Zhang, B., Liu, J., Guo, G., Ji, R.: Bayesian
  optimized 1-bit cnns. In: ICCV (2019)

\bibitem{MoBiNet}
Hai~Phan, Dang~Huynh, Y.H.M.S.Z.S.: Mobinet: A mobile binary network for image
  classification. In: WACV (2020)

\bibitem{He_2017_ICCV}
He, K., Gkioxari, G., Dollar, P., Girshick, R.: Mask r-cnn. In: ICCV (2017)

\bibitem{he2016resnet}
He, K., Zhang, X., Ren, S., Sun, J.: Deep residual learning for image
  recognition. In: CVPR (2016)

\bibitem{Huang_2019_ICCV}
Huang, Z., Wang, X., Huang, L., Huang, C., Wei, Y., Liu, W.: Ccnet: Criss-cross
  attention for semantic segmentation. In: ICCV (2019)

\bibitem{BNN}
Hubara, I., Courbariaux, M., Soudry, D., El-Yaniv, R., Bengio, Y.: Binarized
  neural networks. In: NeurIPS (2016)

\bibitem{ioffe2015BN}
Ioffe, S., Szegedy, C.: Batch normalization: Accelerating deep network training
  by reducing internal covariate shift. In: ICML (2015)

\bibitem{cifar10}
Krizhevsky, A.: Learning multiple layers of features from tiny images  (2009)

\bibitem{krizhevsky2012alexnet}
Krizhevsky, A., Sutskever, I., Hinton, G.E.: Imagenet classification with deep
  convolutional neural networks. In: NeurIPS (2012)

\bibitem{li2016TWN}
Li, F., Zhang, B., Liu, B.: Ternary weight networks. arXiv preprint
  arXiv:1605.04711  (2016)

\bibitem{lin2017ABC-Net}
Lin, X., Zhao, C., Pan, W.: Towards accurate binary convolutional neural
  network. In: NeurIPS (2017)

\bibitem{liu2019auto}
Liu, C., Chen, L.C., Schroff, F., Adam, H., Hua, W., Yuille, A.L., Fei-Fei, L.:
  Auto-deeplab: Hierarchical neural architecture search for semantic image
  segmentation. In: CVPR (2019)

\bibitem{liu2019ganbin}
Liu, C., Ding, W., Xia, X., Hu, Y., Zhang, B., Liu, J., Zhuang, B., Guo, G.:
  Rectified binary convolutional networks for enhancing the performance of
  1-bit dcnns. In: IJCAI (2019)

\bibitem{Liu2019CBCN}
Liu, C., Ding, W., Xia, X., Zhang, B., Gu, J., Liu, J., Ji, R., Doermann, D.:
  Circulant binary convolutional networks: Enhancing the performance of 1-bit
  dcnns with circulant back propagation. In: CVPR (2019)

\bibitem{Liu_2019_ICCV}
Liu, Z., Mu, H., Zhang, X., Guo, Z., Yang, X., Cheng, K.T., Sun, J.:
  Metapruning: Meta learning for automatic neural network channel pruning. In:
  ICCV (2019)

\bibitem{liu2018bireal}
Liu, Z., Wu, B., Luo, W., Yang, X., Liu, W., Cheng, K.T.: Bi-real net:
  Enhancing the performance of 1-bit cnns with improved representational
  capability and advanced training algorithm. In: ECCV (2018)

\bibitem{long2015fully}
Long, J., Shelhamer, E., Darrell, T.: Fully convolutional networks for semantic
  segmentation. In: CVPR (2015)

\bibitem{mishra2017apprentice}
Mishra, A., Marr, D.: Apprentice: Using knowledge distillation techniques to
  improve low-precision network accuracy. arXiv preprint arXiv:1711.05852
  (2017)

\bibitem{polino2018model}
Polino, A., Pascanu, R., Alistarh, D.: Model compression via distillation and
  quantization. arXiv preprint arXiv:1802.05668  (2018)

\bibitem{rastegari2016xnor}
Rastegari, M., Ordonez, V., Redmon, J., Farhadi, A.: Xnor-net: Imagenet
  classification using binary convolutional neural networks. In: ECCV (2016)

\bibitem{VGG}
Simonyan, K., Zisserman, A.: Very deep convolutional networks for large-scale
  image recognition. In: ICLR (2014)

\bibitem{Wan2018TBN}
Wan, D., Shen, F., Liu, L., Zhu, F., Qin, J., Shao, L., Tao~Shen, H.: Tbn:
  Convolutional neural network with ternary inputs and binary weights. In: ECCV
  (2018)

\bibitem{wang2019CI-BCNN}
Wang, Z., Lu, J., Tao, C., Zhou, J., Tian, Q.: Learning channel-wise
  interactions for binary convolutional neural networks. In: CVPR (2019)

\bibitem{Zagoruyko2017AT}
Zagoruyko, S., Komodakis, N.: Paying more attention to attention: Improving the
  performance of convolutional neural networks via attention transfer. In: ICLR
  (2017)

\bibitem{zhang2018shufflenet}
Zhang, X., Zhou, X., Lin, M., Sun, J.: Shufflenet: An extremely efficient
  convolutional neural network for mobile devices. In: CVPR (2018)

\bibitem{zhou2016dorefa}
Zhou, S., Wu, Y., Ni, Z., Zhou, X., Wen, H., Zou, Y.: Dorefa-net: Training low
  bitwidth convolutional neural networks with low bitwidth gradients. arXiv
  preprint arXiv:1606.06160  (2016)

\bibitem{zhu2019BENN}
Zhu, S., Dong, X., Su, H.: Binary ensemble neural network: More bits per
  network or more networks per bit? In: CVPR (2019)

\bibitem{Zhu_2019_ICCV}
Zhu, Z., Xu, M., Bai, S., Huang, T., Bai, X.: Asymmetric non-local neural
  networks for semantic segmentation. In: ICCV (2019)

\bibitem{zhuang2018towards}
Zhuang, B., Shen, C., Tan, M., Liu, L., Reid, I.: Towards effective
  low-bitwidth convolutional neural networks. In: CVPR (2018)

\bibitem{Zhuang2019SBNN}
Zhuang, B., Shen, C., Tan, M., Liu, L., Reid, I.: Structured binary neural
  networks for accurate image classification and semantic segmentation. In:
  CVPR (2019)

\end{thebibliography}
\end{document}